\documentclass{article}

\PassOptionsToPackage{numbers,sort&compress}{natbib}

\usepackage[preprint]{neurips_2026}

\usepackage[utf8]{inputenc} %
\usepackage[T1]{fontenc}    %
\usepackage{hyperref}       %
\usepackage{url}            %
\usepackage{booktabs}       %
\usepackage{amsfonts}       %
\usepackage{nicefrac}       %
\usepackage{microtype}      %
\usepackage{xcolor}         %
\usepackage[pdftex]{graphicx}
\usepackage{subcaption}
\usepackage[table]{xcolor}
\usepackage{amsmath}
\usepackage{printlen}
\usepackage{colortbl}
\usepackage{siunitx}
\usepackage{array}
\usepackage{caption}

\newcommand{\gc}[1]{\textcolor{gray}{#1}}

\definecolor{baseline}{rgb}{0.85, 0.85, 0.85}
\newcommand{\hl}[1]{\cellcolor{baseline}#1}
\usepackage{tabularx}
\newcolumntype{C}{>{\centering\arraybackslash}X}
\definecolor{deltagreen}{HTML}{2A7F2A}
\definecolor{rowtint}{HTML}{F5F5F5}
\definecolor{categorycolor}{HTML}{555555}
\newcolumntype{R}{>{\raggedleft\arraybackslash}X}

\title{TrAction: Action Recognition with Sparse Trajectories}

\author{%
  Jan F.~Meier$^1$, Felix B. Müller$^1$, Alexander Ecker$^{1,2}$, Timo Lüddecke$^1$ \vspace{4pt}\\
  $^1$ \footnotesize Institute of Computer Science and Campus Institute Data Science, University Göttingen, Germany\\
  $^2$ \footnotesize Max Planck Institute for Dynamics and Self-Organization, Göttingen, Germany\vspace{4pt}\\
  \footnotesize\texttt{jan.meier@cs.uni-goettingen.de}
}
\definecolor{darkgreen}{rgb}{0.0, 0.5, 0.0}
\definecolor{darkred}{rgb}{0.5, 0.1, 0.0}

\begin{document}

\maketitle

\begin{abstract}

Modern action recognition models operate on memory- and compute-intensive dense RGB video volumes and frequently exploit appearance and background shortcuts, for example, predicting actions from objects or scenes instead of characteristic motion. We investigate an efficient alternative input modality that is largely free of such biases by construction: sparse point trajectories.
To this end, we develop a simple transformer architecture for 2.5D trajectory-based recognition together with a masked-trajectory pretraining, which we show to substantially improve downstream action recognition accuracy. 
Despite using only a fraction of the dense RGB input, our method reaches 45\% top-1 on Something-Something V2 and 54\% on EPIC-Kitchens-100, and surpasses V-JEPA on time-reversal sensitivity.
More importantly, we find trajectory features to be complementary to state-of-the-art appearance-based features. Fusing our pretrained model with DINOv2 and V-JEPA 2 improves top-1 accuracy on Something-Something V2 by 8.7 and 1.6 points, respectively. 
\\
Code: \url{https://github.com/ecker-lab/TrAction}
\end{abstract}

\section{Introduction}
\label{sec:intro}

\begin{figure}[h!]
    \centering
    \includegraphics[width=\textwidth]{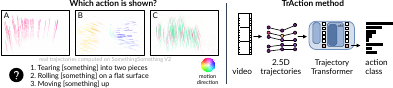}
    \caption{Motion trajectories obtained from CoTracker3 are a sparse yet expressive video representation. Left: Can you guess the actions? Right: We propose a transformer-based neural network and a pretraining method to obtain competitive results in action classification based on trajectories.}
    \label{fig:multi}
\end{figure}

\footnotetext[1]{Solution of Figure 1: 1B, 2C, 3A}

Large-scale video models such as V-JEPA 2~\cite{assran_vjepa_2025} and VideoPrism~\cite{zhao_videoprism_2024} are commonly used for vision tasks like action recognition, as they promise to capture temporal dynamics by jointly processing several dense video frames. In practice, this regime suffers from two fundamental limitations. First, memory and compute scale poorly with resolution and clip length, making long or high-resolution video intractable. Second, the richness of dense pixel inputs invites spurious correlations and the well-documented appearance bias~\cite{sun_masked_2023, brookes_panaffgbg_2025, wang_removing_2021, choi_why_2019}: video models excel at recognizing \emph{what} appears in a video, but often lack sensitivity to the \emph{temporal dynamics of actions}. The advent of accurate dynamic point trackers enables a different novel modality on which action recognition can be based: sparse, motion-only point trajectories as a representation that sidesteps both dense volumes and strong appearance biases by construction.

Motion-focused representations have a long history in action recognition, most prominently through optical flow~\cite{sevilla-lara_integration_2018}. Yet flow is a \emph{dense} per-pixel signal, thus showing similar limitations to RGB frames: poor computational scaling behavior with input size and appearance bias. While it suppresses texture, it leaks object shape, contour and layout information through the spatial extent of moving regions~\cite{ilic_appearance_2022}. Sparse point trajectories, which track multiple 2D points across frames throughout a video, are a more compact representation with a substantially weaker appearance signal: A few hundred randomly sampled point trajectories contain limited information about shape, but suffice to discriminate many motion-heavy actions (see \autoref{fig:multi}). 
This makes them a promising candidate for building an explicit representation of temporal dynamics, complementary to standard vision encoders.
Reliable extraction of such point trajectories has become possible only recently with modern dynamic point trackers~\cite{karaev_cotracker3_2025, zholus_tapnext_2025} producing point trajectories for long sequences while maintaining tracks through occlusions. 
While point trajectories have received attention across multiple disciplines, including robotics and video generation \cite{bharadhwaj_track2act_2024, werby_articulated_2025, chu_wanmove_2026}, their application for action recognition is underexplored so far. 

We address this gap by training a transformer-based action recognition model that operates directly on point trajectories. As 2D trajectories lack important information about scene depth, we augment the 2D point trajectories with depth obtained from a monocular depth estimation, yielding 2.5D trajectories. This approach achieves surprisingly good results on the motion-heavy datasets Something-Something v2 (SSv2) and EPIC-Kitchens-100 (EK100) – despite input sparsity and novelty of the paradigm. As state-of-the-art vision models are largely appearance-focused, a natural next question is, whether motion trajectories are complementary to vision model features. We find that they are: Fusing both modalities yields steady gains across SSv2 and EK100, with the largest improvements on motion-discriminative subsets where appearance models are weakest. We further show that masked-trajectory pretraining improves the trajectory model substantially, analogous to masked auto-encoding in the pixel domain. We carefully ablate model architecture, trajectory sampling and pretraining to help this emerging modality gain TrAction.

In summary, our contributions are:
\setlength{\leftmargini}{2em} 
\begin{itemize}
    \item We propose a method for action recognition from point trajectories, comprising 2.5D trajectory extraction (augmenting CoTracker3 trajectories with monocular depth) and a trajectory transformer.
    \item We introduce a self-supervised masked-trajectory pretraining objective that consistently improves downstream classification (around 5 points top-1 on SSv2 and EK100) and transfers across varying pretraining/evaluation dataset combinations.
    \item We systematically study sparse motion trajectories as an input modality for action recognition. We find trajectories to be complementary to features of vision encoders such as DINOv2 and V-JEPA 2 and show that they can even outperform V-JEPA in special scenarios (Chirality-in-Action benchmark). 
\end{itemize}

\section{Related work}
\label{sec:related}

\paragraph{Action recognition.}
Video action recognition assigns an action label to a video clip and is a central problem in video understanding, evaluated on benchmarks ranging from web videos~\cite{carreira_quo_2017} to object-centric interactions~\cite{goyal_something_2017} and egocentric recordings~\cite{damen_rescaling_2022}. Early deep learning approaches applied image-level CNNs to individual frames~\cite{karpathy_largescale_2014}, followed by recurrent models for temporal reasoning~\cite{donahue_longterm_2015,ballas_delving_2016,yue-heing_short_2015}. Spatiotemporal convolutions~\cite{tran_learning_2015, feichtenhofer_slowfast_2019, feichtenhofer_x3d_2020} introduced joint space-time feature learning directly from RGB, and video transformers subsequently replaced convolutions with self-attention~\cite{bertasius_spacetime_2021, arnab_vivit_2021}. The current dominant paradigm pretrains video transformers with self-supervised objectives such as masked autoencoding~\cite{tong_videomae_2022, wang_videomae_2023, feichtenhofer_masked_2022, zoran_recurrent_2025} or feature prediction~\cite{bardes_revisiting_2024, assran_vjepa_2025} on large-scale unlabeled video.
 
A parallel line of work has explored explicit \emph{motion representations} beyond RGB. Two-stream architectures~\cite{simonyan_twostream_2014, feichtenhofer_convolutional_2016} augment an appearance branch with a dedicated optical flow stream, decoupling motion from appearance; I3D~\cite{carreira_quo_2017} extends this to spatiotemporal convolutions with a two-stream design, and~\cite{sevilla-lara_integration_2018} provides a systematic study of how flow integration affects recognition. Optical flow has also been adopted as a training signal for self-supervised representation learning~\cite{han_memoryaugmented_2020, wang_poodle_2025}. Skeleton-based methods model human pose dynamics through graph convolutions~\cite{yan_spatial_2018, li_cooccurrence_2018, shi_skeletonbased_2019} or attention mechanisms~\cite{song_endtoend_2017, liu_global_2017}, offering a structured, appearance-invariant motion representation for scenarios where reliable pose estimation is available, though limited to human body motion and reliant on domain-specific detectors. \citet{xue_seeing_2025} demonstrated that camera trajectories alone carry significant scene-level information for recognition. Recently, point trajectories have begun to emerge as a recognition modality: TrackMAE~\cite{vandeghen_trackmae_2026} uses them as a self-supervised pretraining signal, while other methods~\cite{kumar_trajectoryaligned_2024, kumar_trokens_2025} combine them with DINO features for few-shot classification. Work concurrent to ours~\cite{davison_its_2026} studies purely trajectories across diverse perceptual tasks, including action recognition. As a general-purpose motion abstraction, point trajectories capture per-point dynamics for arbitrary scene elements without requiring a domain-specific detector, naturally extending both the local, frame-pair scope of optical flow and the sparse, human-centric structure of skeleton representations.

\paragraph{Dynamic point tracking.}
Optical flow, the task of predicting dense per-pixel correspondence between consecutive frames, is a foundational motion estimation problem, addressed by modern methods such as RAFT~\cite{teed_raft_2020} and WAFT~\cite{wang_waft_2026}. While optical flow provides dense frame-pair motion, it does not directly provide long-range point identity or explicit occlusion reasoning across clips. Dynamic point tracking, formalized as Track Any Point (TAP) by~\cite{doersch_tapvid_2022}, addresses this. Given a set of query points $(x, y, t)$, the task is to predict their positions and visibility across all frames. TAP-Net~\cite{doersch_tapvid_2022}, introduced alongside this formalization, applies cost volumes to match query features across frames independently. Subsequent methods improved tracking through iterative refinement~\cite{harley_particle_2022}, coarse-to-fine pipelines combining cost volumes with refinement~\cite{doersch_tapir_2023}, and joint multi-point tracking via inter-point attention~\cite{karaev_cotracker_2024}, which significantly improves robustness under occlusion. More recent work has focused on efficiency, chaining optical flow estimates with lightweight refinement~\cite{lemoing_dense_2024} or restricting cost volumes to local neighborhoods~\cite{cho_local_2024}, as well as on reformulating tracking as sequential token prediction~\cite{zholus_tapnext_2025,jung_tapnext_2026}. On the training side, CoTracker3~\cite{karaev_cotracker3_2025} and BootsTAP~\cite{doersch_bootstap_2024} address the synthetic-to-real domain gap by generating pseudo-labels from existing trackers on unlabeled video. A parallel direction lifts point tracking into 3D, using geometric priors~\cite{xiao_spatialtracker_2024, xiao_spatialtrackerv2_2025} or factored spatiotemporal attention~\cite{ngo_delta_2025} to predict 3D trajectories.

\paragraph{Point trajectories as a motion representation.}
Beyond tracking itself, point trajectories have rapidly become a versatile representation adopted across diverse vision tasks. They serve as compact state representations for robotic manipulation~\cite{zhang_dreamvla_2026, werby_articulated_2025, bharadhwaj_track2act_2024, chen_historyaware_2025}, as user-controllable motion handles for video generation~\cite{lee_generative_2025, chu_wanmove_2026}, and as geometric priors for 3D reconstruction~\cite{wang_vggsfm_2024,wang_shape_2025,yin_trackersplat_2025}. Point tracks have further been applied to video object segmentation~\cite{rajic_segment_2025}, multi-object tracking~\cite{zheng_nettrack_2024}, and motion forecasting~\cite{thakkar_forecasting_2026}.

For action recognition specifically, concurrent work has begun exploring trajectory-based representations: TRec~\cite{holzmann_trec_2026} fuses point tracks with image features via early fusion, and the methods discussed above~\cite{vandeghen_trackmae_2026, kumar_trajectoryaligned_2024, kumar_trokens_2025, davison_its_2026} each incorporate trajectories in different settings. However, these approaches either rely on hybrid trajectory-appearance input~\cite{holzmann_trec_2026, kumar_trajectoryaligned_2024, kumar_trokens_2025}, are limited to few-shot evaluation~\cite{kumar_trajectoryaligned_2024, kumar_trokens_2025}, or address action recognition only as a secondary task~\cite{davison_its_2026}. We systematically evaluate trajectories on standard action recognition benchmarks, analyze the information they carry, and demonstrate their complementarity with state-of-the-art vision models.

\section{Method: Trajectories for action recognition (TrAction) }
\label{sec:methods}

\begin{figure}[tb]
    \centering
    \includegraphics[width=\linewidth]{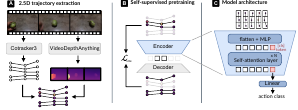}
    \caption{\textbf{Trajectories for action recognition (TrAction) overview.}
    We extract 2.5D trajectories using Cotracker3 and VideoDepthAnything (A). Our trajectory transformer model is first pretrained using self-supervised masked autoencoding (B) before being finetuned for action recognition (C).}
    \label{fig:overview}
\end{figure}

We investigate how much semantic information point trajectories carry for action recognition and how to access it effectively. Our model operates \emph{exclusively} on point trajectories and receives no RGB frames, optical flow, or appearance features as input. We design a simple yet effective architecture inspired by vision transformers, paired with a pipeline for extracting 2.5D trajectories from video, suitable tokenization, normalization, and a self-supervised pretraining strategy. \autoref{fig:overview} illustrates the full pipeline, organized into three stages: trajectory extraction, model architecture, and self-supervised pretraining.

\paragraph{Trajectory extraction.}
The dynamic point tracker takes $N$ query points $(u, v, t)$ describing point locations in pixel coordinates $(u, v)$ and time $t$. We use CoTracker3~\cite{karaev_cotracker3_2025} in offline mode, which propagates points forward and backward through the entire clip of length $T$. Each trajectory is represented as a sequence of 2D positions and a visibility score: $(u_t, v_t, \mathrm{vis}_t)_{t=1}^{T}$, where $\mathrm{vis}_t \in \left[ 0,1 \right]$ indicates whether the point is visible at frame~$t$. We sample query points uniformly at random across both space and time. Querying at a single frame would systematically miss objects entering the scene mid-clip, which is common in action videos where the action's onset rarely coincides with frame zero. 

To lift trajectories from 2D to 2.5D, we augment each point with a depth value obtained from Video Depth Anything~\cite{chen_video_2025} at position $(u_t, v_t)$ in frame~$t$, yielding $(u_t, v_t, d_t, \mathrm{vis}_t)$. We opt for this modular design, combining a 2D tracker with a separate depth estimator, rather than using an end-to-end 3D tracker. This provides flexibility in model selection and leverages the relative maturity and efficiency of 2D tracking methods compared to current 3D tracking approaches, which typically require costly scene reconstruction~\cite{xiao_spatialtrackerv2_2025}.

\paragraph{Model architecture.}
Each trajectory $\mathbf{p} \in \mathbb{R}^{T \times 4}$ is parameterized as $(u, v, d, \mathrm{vis})$ per frame. We normalize the spatial coordinates $u, v$ by dividing by $\max(H, W)$, computed per sample, where $H$ and $W$ are the video height and width in pixels. Compared to independent normalization to image coordinates (dividing $u$ by $W$ and $v$ by $H$), this preserves the aspect ratio and spatial structure of the scene, ensuring that the relative shapes and proportions of object motions remain consistent. The depth channel $d$ is normalized per sample to have zero mean and the same standard deviation as the spatial coordinates, placing all geometric channels on a common scale.

We apply sinusoidal frequency encoding to the three geometric channels $(u, v, d)$, transforming each scalar through $L$ frequency bands. This yields $\mathbf{p}_{\mathrm{enc}} \in \mathbb{R}^{T \times (3 \cdot 2L + 1)}$, where the $+1$ accounts for the visibility. The encoded trajectory is flattened across time into a single vector and projected to the model dimension $D_{model}$ via a two-layer MLP with GELU activation $\sigma$ and dropout:

\begin{equation}
\label{eq:proj-head}
    \mathbf{z} = \mathbf{W}_2 \, \sigma\!\left( \operatorname{Dropout}\!\left( \mathbf{W}_1 \, \operatorname{flatten}(\mathbf{p}_{\mathrm{enc}}) + \mathbf{b}_1 \right) \right) + \mathbf{b}_2.
\end{equation}

This yields one token $\mathbf{z} \in \mathbb{R}^{D}$ per trajectory. Each token summarizes the full temporal extent of a single query point: temporal reasoning is handled within the per-trajectory MLP, while spatial and inter-point reasoning emerges through self-attention in the subsequent transformer layers. A 3D sinusoidal positional encoding based on each trajectory's starting position $(u_0, v_0, d_0)$ is added to the token, providing the model with absolute spatial context. A learnable \texttt{[CLS]} and 4 register tokens are prepended, and the resulting $m{+}5$ tokens are processed by a stack of standard transformer encoder layers. For classification, the \texttt{[CLS]} token after the final layer is passed through a linear projection to the number of classes, trained with cross-entropy loss. Architecture hyperparameters are detailed in \autoref{app:hyperparams}.

\paragraph{Self-supervised pretraining.}
Self-supervised pretraining is essential for strong downstream performance in modern vision models~\cite{tong_videomae_2022, feichtenhofer_masked_2022}. Inspired by this, we pretrain our model with a masked autoencoding (MAE) objective adapted for trajectories. During pretraining, we randomly mask a fraction $p_{\mathrm{mask}}$ of the $m$ trajectory tokens. Following~\cite{tong_videomae_2022}, we use an asymmetric encoder-decoder design: only the unmasked tokens are processed by the full encoder, reducing pretraining cost. The lightweight decoder consists of a bottleneck MLP followed by two transformer layers. Masked tokens are represented as a shared learned embedding with added positional encoding and are introduced only at the decoder stage. The reconstruction target are the raw spatial coordinates $(u, v, d)$ of each masked trajectory across all frames. We minimize the mean absolute error between the predicted and ground-truth coordinates in the normalized coordinate space. After pretraining, the decoder is discarded and only the encoder weights are transferred to the downstream classification task.

\section{Experimental setup}
\label{sec:exp}

\paragraph{Datasets.}
We evaluate action recognition on three standard benchmarks. \textbf{Kinetics-400}~\cite{carreira_quo_2017} is appearance-dominated and included as a reference, while \textbf{Something-Something v2}~\cite{goyal_something_2017} requires fine-grained temporal reasoning. On \textbf{EPIC-Kitchens-100}~\cite{damen_rescaling_2022}, we report only verb prediction, since noun recognition is inherently appearance-driven. We use two additional datasets for more targeted analyses:  \textbf{IARD}~\cite{tacchetti_invariant_2017} to assess motion-appearance disentanglement via $k$-NN classification following~\cite{ressler-antal_dismo_2026} and \textbf{Chirality-in-Action}~\cite{bagad_chirality_2026} to assess discrimination between temporally mirrored actions. More extensive dataset descriptions and statistics are in \autoref{app:datasets}.

\paragraph{Implementation details.}
We extract $M{=}2500$ trajectories per video using CoTracker3-offline~\cite{karaev_cotracker3_2025} and depth from Video Depth Anything-L~\cite{chen_video_2025}. During training we randomly sample $m$ trajectories per forward pass. This reduces compute and serves as a straightforward form of data augmentation, since the model observes a different point set at each iteration and must learn representations that are invariant to the specific subset of points. During evaluation we use all $M$ trajectories as an input. Except for experiments on IARD and Chirality-in-Action, where $m{=}2500$, we consistently use $m{=}512$. Our model consists of six transformer encoder layers with hidden dimension $D_{\text{model}}{=}288$, totaling less than 8M parameters. Trajectory augmentations and remaining hyperparameters are listed in \autoref{app:augments} and \ref{app:hyperparams}.

For vision-model baselines, we precompute features once per video (16 frames, single clip, single crop) and train an attentive probe~\cite{bardes_revisiting_2024} on the frozen features. The I3D baseline is fully-finetuned. Input resolution follows each model's native setting. For complementary fusion, trajectory and vision features are concatenated and passed through a linear layer trained from scratch, isolating feature complementarity from the choice of fusion architecture. All experiments run on 1-4 A100 or H100 GPUs. Full setups including configurations used for baseline training are described in \autoref{app:detail_exp}.

\section{Results}
\label{sec:results}
\subsection{Trajectories as a modality for action recognition}

\begin{table}[tb]
    \centering
    \small
    \setlength{\tabcolsep}{4pt}
    \caption{\textbf{Trajectory-only recognition and complementarity with vision models.}
             Top-1 accuracy on K400, EK100, and SSv2. ``+ Ours'' columns fuse the
             appearance backbone with our trajectory model via late fusion. We do not use test-time augmentations.
             EK100 results are on val.\\ $^*$Result from \cite{carreira_quo_2017}, includes test-time augmentations and longer sequences. %
             }
    \label{tab:results_main}
    \renewcommand{\arraystretch}{1.15}
    \begin{tabularx}{\textwidth}{l@{\,}r l@{\,}r C C C C C}
        \toprule
         & & & & \multicolumn{5}{c}{\textit{motion-driven \,$\longleftarrow\!\!\!\longrightarrow$\, appearance-driven}} \\
         \cmidrule{5-9}
        \textbf{Model} & & \multicolumn{2}{l}{\textbf{Input size}}
         & \multicolumn{2}{c}{\textbf{SSv2}}
         & \multicolumn{2}{c}{\textbf{EK100}}
         & \textbf{K400} \\
        \midrule
       \textcolor{categorycolor}{\textit{Non-RGB}} \\

        Ours & &\textit{\textit{32\texttimes2500\texttimes 4}} & = 0.3M
            & \multicolumn{2}{c}{45.2} 
            & \multicolumn{2}{c}{54.1} 
            & 33.1 \\
        I3D (flow-only) &\cite{carreira_quo_2017} &\textit{\textit{16\texttimes224$^\textit{2}$\texttimes 2}} & = 1.6M
            & \multicolumn{2}{c}{38.4}
            & \multicolumn{2}{c}{54.0}
            & 63.4\rlap{$^{*}$} \\
        \midrule
        \textcolor{categorycolor}{\textit{Image models}} & & & &
            \textbf{base} & \textbf{+ Ours} & \textbf{base} & \textbf{+ Ours} & \\
        \cmidrule(lr){5-6} \cmidrule(lr){7-8}
        DINOv2-B (center) & \cite{oquab_dinov2_2024} &
        \textit{1\texttimes224$^\textit{2}$\texttimes 3} & = 0.2M
            & 30.8 & 56.8\,{\color{deltagreen}\footnotesize(+26.0)}
            & 44.1 & 59.5\,{\color{deltagreen}\footnotesize(+15.4)}
            & 66.5 \\
        DINOv2-B  & \cite{oquab_dinov2_2024} &
        \textit{16\texttimes224$^\textit{2}$\texttimes 3} & = 2.4M
            & 53.6 & 62.3\,{\color{deltagreen}\footnotesize(+8.7)}
            & 59.1 & 63.5\,{\color{deltagreen}\footnotesize(+4.4)}
            & 73.5 \\
        SigLIP-2-B  & \cite{tschannen_siglip_2025} &
        \textit{16\texttimes224$^\textit{2}$\texttimes 3} & = 2.4M
            & 51.5 & 61.4\,{\color{deltagreen}\footnotesize(+9.9)}
            & 56.6 & 63.4\,{\color{deltagreen}\footnotesize(+6.8)}
            & 73.6 \\
        \midrule
        \textcolor{categorycolor}{\textit{Video models}} & & & &
            \textbf{base} & \textbf{+ Ours} & \textbf{base} & \textbf{+ Ours} & \\
        \cmidrule(lr){5-6} \cmidrule(lr){7-8}
        VideoPrism-B & \cite{zhao_videoprism_2024} & \textit{16\texttimes224$^\textit{2}$\texttimes 3} & = 2.4M
            & 60.5 & 63.5\,{\color{deltagreen}\footnotesize(+3.0)}
            & 62.9 & 64.5\,{\color{deltagreen}\footnotesize(+1.6)}
            & \bfseries 76.6 \\
        V-JEPA 2-L & \cite{assran_vjepa_2025} & \textit{16\texttimes256$^\textit{2}$\texttimes 3} & = 3.1M
            & 66.7 & {\bfseries 68.3}\,{\color{deltagreen}\footnotesize(+1.6)}
            & 67.2 & {\bfseries 68.3}\,{\color{deltagreen}\footnotesize(+1.1)}
            & 73.0 \\
        \bottomrule
    \end{tabularx}
\end{table}

\setlength{\parskip}{0.4em} 

\paragraph{Trajectory-only action recognition works surprisingly well.} 

First, we analyze how well actions can be recognized exclusively from trajectories, without appearance information. In this setting we observe a good performance on motion-driven datasets while performance drops when appearance matters. Trajectories outperform optical flow on the motion-driven SSv2 dataset  (45.2 vs.\ 38.4), while this effect is less pronounced on EK100 (54.1 vs.\ 54.0). We attribute this to more frequent appearance leaks of the actions in EK100: Kitchen verbs are often tied to specific tools and interactions (e.g. cutting is associated with a knife).
Trajectories perform worse than the powerful dense multi-frame DINOv2, SigLIP-2, and V-JEPA 2-L baselines. This is expected as appearance still provides a valuable signal on these datasets.
On K400 this gap between trajectories and vision models grows. This is unsurprising given that K400 is known for is strong appearance bias: many of its action classes are recognizable from a single frame ~\cite{kowal_deeper_2022, li_resound_2018, huang_what_2018}.

\paragraph{Our trajectory model consistently complements vision models.}
Are point trajectories redundant with classic vision models features or do they complement them? To address this question, we fuse our trajectory encoder with multiple image and video models of increasing capacity. For each, we train an attentive probe and combine it with our trajectory encoder via late fusion (\autoref{sec:exp}). Across all vision models, fusion yields consistent gains on both SSv2 and EK100 (\autoref{tab:results_main}, green). Most notably, fusion improves V-JEPA 2-L by $+1.6$ on SSv2 and $+1.1$ on EK100, despite V-JEPA 2-L being one of the strongest self-supervised video encoders available. Trajectories increasing accuracy of such a model suggests that RGB pretraining does not internalize the full motion structure of actions, and that sparse trajectories carry a recognition signal inaccessible  to the dense video and image encoders. In general, gains are most pronounced on image models which are less able to capture motion information themselves. Combining center-frame-only DINOv2 with our model outperforms DINOv2 on 16 frames with a fifth of input size. The complementary gain decreases as the base model strengthens, as stronger video models recover more motion information from RGB on their own.

\begin{figure}[t]
    \centering
    \begin{subfigure}[t]{0.48\textwidth}
        \centering
        \includegraphics[width=\textwidth]{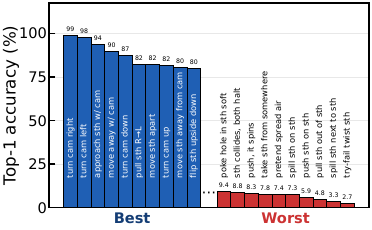}
        \caption{\textbf{Best and worst performing classes for trajectories only.} Trajectories perform well on motion-only classes, but struggle when appearance information is needed.}
        \label{fig:frames}
    \end{subfigure}
    \hfill
    \begin{subfigure}[t]{0.48\textwidth}
        \centering
        \includegraphics[width=\textwidth]{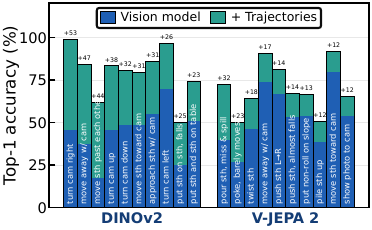}
        \caption{\textbf{Largest performance gains when combining vision models with trajectories.} Performance improvement over the vision model base are visualized as green bars on top.}
        \label{fig:trajectories}
    \end{subfigure}
    \caption{\textbf{Class-wise performance on SSv2.} (a) The trajectories only model performs well on actions involving camera motion as well as directional classes. (b) Fusing both DINOv2 as well as V-JEPA 2 with our trajectory model leads to significant gains. Classes with less than 25 samples are excluded and class labels are shortened.}
    \label{fig:classes_vis}
    \vspace{-0.5cm}
\end{figure}

\paragraph{Not all classes benefit equally from trajectories.}
Class-wise performances (\autoref{fig:classes_vis}) reveal that individual classes involving distinct, motion patterns often benefit substantially, while classes whose action label depends on object identity or scene context (\textit{push sth. on sth.}, \textit{pretend to spread air on sth.}) do not benefit or drop accuracy. 
Considering the classes with the highest gains on DINOv2, we see that trajectories primarily contribute information about camera motion. On V-JEPA 2, which likely encodes camera motion reasonably well, the gains are smaller and mostly affect object motions.
Taken together, these findings suggest that trajectories indeed compensate for the lack of motion of the base features and are thus complementary.

\paragraph{Our model exhibits sensible attention patterns}
\begin{figure}[tb]
    \centering
    \includegraphics[width=\linewidth]{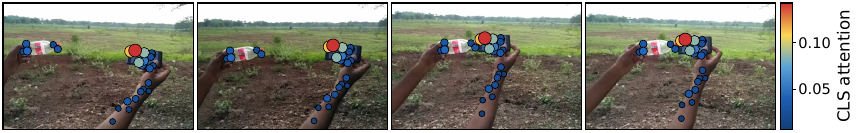}
    \caption{\textbf{Last-layer CLS attention overlay.} Top-25 trajectories from one of four heads on a \textit{moving X closer to Y} sequence. Color encodes attention weight, size scales with weight, alpha with trajectory visibility. The visualized head concentrates on the manipulated object; other heads attend to different regions and motions. See \autoref{app:vis} for additional examples.}
    \label{fig:attention_vis}
    \vspace{-0.5cm}
\end{figure}

\autoref{fig:attention_vis} visualizes last-layer CLS attention on a \textit{moving X closer to Y} sequence. The visualized head is highly selective: 90\% of its attention falls on 5\% of trajectories, clustered on the manipulated object rather than the moving hand or arm and tracking the object across frames. The remaining heads attend to different regions and motions. This qualitative view is illustrative; we do not draw quantitative claims from it. \autoref{app:vis} shows additional examples across categories.

\subsection{Why the trajectory modality complements vision models} 
The complementarity between trajectory and vision models shown in \autoref{tab:results_main} raises the question of what kind of information trajectories provide. We argue that trajectories are by construction largely invariant to appearance and aware of temporal dynamics, properties that pretrained appearance models acquire only partially and only at scale. We support this through two diagnostic benchmarks: IARD~\cite{tacchetti_invariant_2017} for motion-appearance disentanglement and Chirality-in-Action~\cite{bagad_chirality_2026} for time-awareness. 

\begin{table}[tb]
    \centering
    \caption{\textbf{Structural properties of the point trajectories as modality.}}
    \label{tab:properties}
    \footnotesize
    \renewcommand{\arraystretch}{1.15}
    \begin{subtable}[t]{0.48\textwidth}
        \centering
        \caption{\textbf{Appearance-invariance.} On IARD~\cite{tacchetti_invariant_2017}, identity is highly recoverable from RGB-pretrained features, whereas our trajectory model is nearly appearance-invariant by construction. Baseline results from \cite{tacchetti_invariant_2017}.}
        \label{tab:properties_iard}
        \begin{tabularx}{\textwidth}{l@{\,}r l R R}
            \toprule
            \textbf{Model} & &\textbf{Pretrain.} & \textbf{Action} & \textbf{Identity} \\
            \midrule
            Random       & &{--}                         & 20.0   & 20.0 \\
            \midrule
            DINOv2-B   &  \cite{oquab_dinov2_2024}   & LVD-142M                  & 66.0 & 99.0 \\
            VideoMAE-L   & \cite{tong_videomae_2022}    & K400                      & 73.4 & 99.1 \\
            V-JEPA-L     & \cite{bardes_revisiting_2024}    & VideoMix2M                & \textbf{82.0} & 96.2 \\
            \gc{DisMo-B} &\cite{ressler-antal_dismo_2026}     & \gc{2.8M clips}     & \gc{90.7} & \gc{23.8} \\
            \midrule
            Ours           & & K400, SSv2                & 76.6 & \textbf{27.3} \\
            \bottomrule
        \end{tabularx}
    \end{subtable}%
    \hfill
    \begin{subtable}[t]{0.48\textwidth}
        \centering
        \caption{\textbf{Time-reversal sensitivity.} On Chirality-in-Action (SSv2 split)~\cite{bagad_chirality_2026}, our model distinguishes forward from reverse actions better than large pretrained video models do. Baseline results  from \cite{bagad_chirality_2026}.}
        \label{tab:properties_chirality}
        \begin{tabularx}{\textwidth}{l@{\,}r l  R}
            \toprule
            \textbf{Model} & &\textbf{Pretrain.} & \textbf{SSv2} \\
            \midrule
            Random       & &{--}                 & 50.0 \\
            \midrule
            DINOv2-S      &  \cite{oquab_dinov2_2024}              & LVD-142M      &  79.7 \\
            SigLIP 2-B    & \cite{tschannen_siglip_2025}              & WebLI         &  76.8 \\
            VideoMAE-L    & \cite{tong_videomae_2022}              & K400          &  85.7 \\
            V-JEPA-L      & \cite{bardes_revisiting_2024}              & VideoMix2M    &  85.4 \\
            \midrule
            Ours                        & & K400, SSv2    &  \textbf{86.2} \\
            \bottomrule
        \end{tabularx}
    \end{subtable}
    \vspace{-0.3cm}
\end{table}
 
\paragraph{Trajectories have low appearance bias.} 

An ideal motion representation would have a high sensitivity for actions while being invariant to appearance cues. We quantify the degree to which this property holds for our trajectories and standard vision models using the IARD ~\cite{tacchetti_invariant_2017} benchmark \autoref{tab:properties}(a) which measures how well action and personal identity labels are preserved in model features using a kNN protocol.
Across all RGB-pretrained models, identity is highly recoverable from features, reflecting the appearance information their pretraining preserves. As a reference point, DisMo~\cite{ressler-antal_dismo_2026}, a foundation-scale model explicitly trained to suppress appearance, reaches 23.8\% identity accuracy, close to the 20.0\% random baseline. Our trajectory model attains a comparable 27.3\% without any appearance-action-disentanglement objective and with a magnitude less pretraining data, while remaining competitive on action recognition. We attribute this to the inherent properties of trajectories as a modality.

\paragraph{Trajectories contain strong time-aware motion information.} 

Some action classes have similar appearance with the main difference being temporal direction, e.g. opening/closing a door. A model's ability to correctly discriminate these chiral classes thus measures (this aspect of) its time-awareness. To evaluate this (\autoref{tab:properties}(b)), we use the recently proposed Chirality-in-Action benchmark by \citet{bagad_chirality_2026}.  
Our trajectory model surpasses state-of-the-art video models on this task, using only point trajectories as a sparse signal that discards appearance entirely. This indicates that temporal information is encoded well in point trajectories.

These two findings suggest a structural source of the complementarity between trajectory and vision models: Trajectories posses an appearance-invariance and time-awareness that vision models acquire only partially, even at scale. Combining the two representations is therefore not a redundant compounding of similar features but an addition of structurally different information.

\subsection{Ablations}

For all ablations, we train our model from scratch without self-supervised pretraining.

\paragraph{Scaling behavior of frame and trajectory counts}

\begin{figure}[t]
    \centering
    \begin{minipage}[t]{0.58\textwidth}
    \vspace*{0pt}
    \centering
    \includegraphics[width=\textwidth]{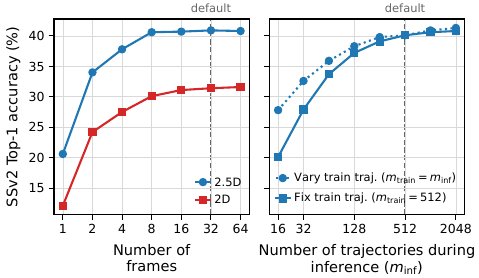}
    \caption{\textbf{Effect of frames and trajectories on SSv2.} Performance increases with more frames and saturates beyond 16. Increasing the number of trajectories helps consistently but gains are small beyond 256 trajectories. }
    \label{fig:ablation_num}
     \end{minipage}
    \hfill
    \begin{minipage}[t]{0.39\textwidth}
    \vspace*{0pt}
    \centering
    \captionof{table}{\textbf{Comparison of trajectories (Ours) and optical flow (I3D) as complementary input to vision models.}
            Trajectories excel on the motion-heavy SSv2 dataset and perform close to optical flow on EK100, despite being $5\times$ smaller.}
        \label{tab:compl_of}
    \footnotesize
            \renewcommand{\arraystretch}{1.1}
            \setlength{\tabcolsep}{4pt}  %
    \begin{tabularx}{\textwidth}{llll}
            \toprule
            \textbf{model}  & \textbf{input} & \textbf{SSv2} & \textbf{EK100} \\
             & \textbf{size} && \\
            \midrule
            DINOv2-B & 2.4M                         
            &  53.6 
            & 59.1 \\
            \textit{+ I3D} & 1.6M                    
            &  60.1\,{\color{deltagreen}\footnotesize(+6.5)} 
            & \textbf{64.1}\,{\color{deltagreen}\footnotesize(+5.0)} \\
            \textit{\textbf{+  Ours}} & 0.3M    
            &  \textbf{62.3}\,{\color{deltagreen}\footnotesize(+8.7)} 
            & 63.5\,{\color{deltagreen}\footnotesize(+4.4)} \\
            \midrule
            V-JEPA-L & 3.1M                         
            &  66.7 
            & 67.2 \\
            \textit{+ I3D} & 1.6M                  
            &  67.9\,{\color{deltagreen}\footnotesize(+1.2)} 
            & \textbf{69.1}\,{\color{deltagreen}\footnotesize(+1.9)} \\
            \textit{\textbf{+ Ours}} & 0.3M         
            &  \textbf{68.3}\,{\color{deltagreen}\footnotesize(+1.6)} 
            & 68.3\,{\color{deltagreen}\footnotesize(+1.1)} \\
            \bottomrule
        \end{tabularx}
        \label{tab:optical_flow}
            \end{minipage}
\end{figure}

We explore how the number of frames and trajectories affects trajectory-only action recognition (\autoref{fig:ablation_num}). First, we vary the number of frames provided during training and inference by linearly interpolating each full-length trajectory to the target length $T$. Unlike re-running tracking at lower frame rates, this subsampling approach decouples the model's frame sensitivity from tracker-side degradation under large strides. Frame-count gains saturate beyond 8 frames, indicating that coarse temporal sampling already captures most of the action-discriminative motion. We also compare 2D point trajectories (without depth information) to our default 2.5D trajectories and that depth information brings a substantial performance gain that is independent of the number of frames.

Second, we investigate how the number of trajectories affects the model. We find that our model performs well with few trajectories. Halving the trajectory count from 512 to 256 during training only costs 0.3 percentage points and even 16 trajectories achieve non-trivial performance (27.8\,\%). Increasing trajectory counts beyond 512 still increases performance albeit at a small rate. Even when trained on a fixed number of trajectories (here: $m_{\text{train}}=512$), our model can take varying numbers of trajectories during inference. Increasing the number of trajectories boosts performance slightly and it still performs well with less trajectories. Highest performance, however, is always achieved when using the same number of trajectories for training and inference. Note that default configuration used throughout the paper is $m_{\text{train}}=512$ and $m_{\text{inf}}=2500$.

\begin{table}[t]
\centering
\footnotesize
\setlength{\tabcolsep}{4pt}
\renewcommand{\arraystretch}{1.1}

\begin{subtable}[t]{0.31\linewidth}
\centering
\begin{tabularx}{\linewidth}{lCC}
\toprule
dataset & SSv2 & EK100 \\
\midrule
no      & 40.9                  & 49.4 \\
K400    & \textbf{45.8}         & 53.9 \\
SSv2    & 45.6                  & 53.8 \\
both    & \hl{\textbf{45.8}}    & \hl{\textbf{54.5}} \\
\bottomrule
\end{tabularx}
\caption{\textbf{Pre-training dataset.} Source dataset for MAE pretraining.}
\label{tab:abl-pretrain-dataset}
\end{subtable}\hspace{0.025\linewidth}%
\begin{subtable}[t]{0.31\linewidth}
\centering
\begin{tabularx}{\linewidth}{lCC}
\toprule
mask ratio & SSv2 & EK100 \\
\midrule
0.60 & 45.2                 & \textbf{54.1} \\
0.75 & 45.6                 & \textbf{54.1} \\
0.90 & \hl{\textbf{45.8}}   & \hl{53.9} \\
0.95 & 45.1                 & 52.6 \\
\bottomrule
\end{tabularx}
\caption{\textbf{Mask ratio}  during pretraining on K400. }
\label{tab:abl-mask-ratio}
\end{subtable}\hspace{0.025\linewidth}%
\begin{subtable}[t]{0.31\linewidth}
\centering
\begin{tabularx}{\linewidth}{lCC}
\toprule
model & SSv2 & EK100 \\
\midrule
base                & \hl{\textbf{40.9}}    & \hl{\textbf{49.4}} \\
w/o sine-enc.       & 40.7                  & 48.5 \\
w/o pos enc.        & 38.4                  & 48.1 \\
rel. coords         & 32.0                  & 42.1 \\
\bottomrule
\end{tabularx}
\caption{\textbf{Encoding of input trajectories.}}
\label{tab:abl-trajectory-model}
\end{subtable}
\vspace{1em}
\begin{subtable}[t]{0.31\linewidth}
\centering
\begin{tabularx}{\linewidth}{lCC}
\toprule
modality & SSv2 & EK100 \\
\midrule
base                    & \hl{\textbf{40.9}}  & \hl{\textbf{49.4}} \\
w/o depth               & 31.4                & 43.5 \\
w/o visibility          & 39.3                & 48.6 \\
\bottomrule
\end{tabularx}
\caption{\textbf{Modalities.} Modality ablation; ``base'' is full model.}
\label{tab:abl-architecture}
\end{subtable}\hspace{0.025\linewidth}%
\begin{subtable}[t]{0.31\linewidth}
\centering
\begin{tabularx}{\linewidth}{lCC}
\toprule
strategy & SSv2 & EK100 \\
\midrule
rand.\ 512              & 40.0                  & 48.7  \\
rand.\ 512$\times$5     & 40.8                  & 49.3  \\
all                     & \hl{\textbf{40.9}}    & \hl{\textbf{49.4}}  \\
\bottomrule
\end{tabularx}
\caption{\textbf{Inference sampling.} Trajectory selection at inference.}
\label{tab:abl-inference-sampling}
\end{subtable}
\vspace{-0.3cm}
\caption{\textbf{Ablation experiments.}
All models share the same backbone and training schedule. The default choice for our model is highlighted in gray \fcolorbox{black}{baseline}{\rule{0pt}{0.3em}\hspace{0.3em}}. Ablations (c) - (e) are performed without pretraining to rule out any influence of the pretraining strategy. Scores are computed on val. set.}
\label{tab:ablations}
\vspace{-0.5cm}
\end{table}

\paragraph{Comparing point trajectories and optical flow as complementary modalities}
As optical flow is also designed to primarily capture motion information, it is a natural comparison to point trajectories. We compare our model to I3D using the same late-fusion approach and find that both modalities consistently complement vision models. Both approaches perform comparable, despite I3D processing dense features $5\times$ the size of point trajectories. Ours outperforms I3D on SSv2, while I3D performs better on EK100, see Table~\ref{tab:optical_flow}. This is consistent with SSv2 and point trajectories being less appearance-biased compared to EK100 and optical flow. 

\paragraph{Design choices}
During the development of our model and pretraining we made several design choices that we assess next (\autoref{tab:ablations}).
\textbf{(a)} Pretraining significantly boosts the accuracy by 4.9 and 5.1 points on SSv2 and EK100 respectively. There is no large difference between different pretraining datasets, which indicates that masked-trajectory pretraining captures generic motion structure rather than dataset-specific pattern. \textbf{(b)} There is no clear trend towards higher or lower masking ratio, except that 0.95 is too high. \textbf{(c)} Passing the three geometric channels $(u, v, d)$ directly (\emph{w/o sine-enc.}) instead of performing sinusoidal frequency encoding slightly reduces performance. Removing the 3D positional encoding from the trajectory tokens (\emph{w/o pos enc.}) negatively impacts the performance, even though the tokenizer has access to the absolute trajectory positions. Surprisingly, only feeding relative trajectories (all coordinates start at the origin in the first frame, \emph{rel. coords}) in the model, while introducing the absolute position only via the positional encoding, severely hurts the performance. \textbf{(d)} The additional depth information boosts the accuracy by 9.5 and 5.9 points on SSv2 and EK100, respectively. Using the visibility of the trajectories as an additional feature also increases the accuracy, but only to a smaller extent. \textbf{(e)} During training, we randomly sample 512 from the 2500 extracted trajectories. During inference, the best approach is to use all 2500 trajectories. This performs better than sampling 5 times 512 trajectories and averaging the logits as a form of test-time augmentation. We use the highlighted configuration as our default throughout the paper.

\subsection{Limitations}
Our approach relies on a separate, computationally expensive trajectory extraction step. This is a one-time cost that is amortized over many training runs by caching trajectories. The  extraction step itself imposes a second limitation: modern point trackers can struggle on highly dynamic scenarios with rapid camera motion or abrupt scene changes. Lastly, motion trajectories are only beneficial when motion is informative for the task. On benchmarks dominated by static scene appearance (e.g., Kinetics-400, UCF-101, HMDB-51), trajectories are not informative.

\section{Conclusion}
\label{sec:conclusion}
We show that sparse point trajectories are a viable modality for action recognition. Despite carrying no pixel-level appearance, a compact transformer trained on trajectories reaches 45\% top-1 on SSv2 and 54\% on EK100. A detailed analysis shows that our approach is competitive with significantly larger models on appearance-invariance (IARD) and time-awareness (Chirality-in-Action). Most importantly, trajectory features are complementary to RGB-based representations: late fusion with frozen DINOv2 or V-JEPA 2 encoders yields consistent gains over each modality alone, with the largest improvements on classes where motion disambiguates appearance. We further show that masked-trajectory pretraining significantly improves downstream performance. While most contemporary video models entangle appearance and motion in a single learned representation, sparse trajectories offer an alternative: an explicit motion signal that can be analyzed, fused, or replaced independently of appearance. 
Trajectories are not a replacement for appearance-based video understanding and fail predictably where appearance dominates. 
But in scenarios where motion carries the signal, point trajectories have proven to be a useful sparse representation of the video.
We hope this motivates further work on explicit motion modalities and their fusion with appearance-based video models.

\section*{Acknowledgments}
The project was funded by the Deutsche Forschungsgemeinschaft (DFG, German Research Foundation) – Project-ID 454648639 – SFB 1528. The authors gratefully acknowledge the computing time granted by the Resource Allocation Board and provided on the supercomputer Emmy/Grete at NHR-Nord@G\"ottingen as part of the NHR infrastructure. The calculations for this research were conducted with computing resources under the project nib00021.

\bibliographystyle{unsrtnat}
\bibliography{literature}

\newpage
\appendix

\section{Datasets}
\label{app:datasets}

We evaluate our method on five action recognition datasets covering a range of video domains, label granularities, and motion characteristics. Table~\ref{tab:datasets} summarizes the key statistics for each dataset.

\paragraph{Kinetics-400 (K400) \cite{carreira_quo_2017}.}
Kinetics-400 is a large-scale action recognition benchmark consisting of short clips collected from YouTube and labeled with 400 human action classes covering a wide range of everyday activities. Clips are roughly 10 seconds long and depict a diverse set of scenes, viewpoints, and actors.

\paragraph{EPIC-Kitchens-100 (EK100) \cite{damen_rescaling_2022}.}
EPIC-Kitchens-100 is a large-scale egocentric video dataset of unscripted cooking activities recorded with head-mounted cameras in real kitchen environments. Each segment is annotated with verb and noun labels. In this work, we evaluate on the 97 verb classes, which capture the type of action being performed (e.g., ``cut'', ``open'', ``pour''). The dataset features substantial camera motion, frequent hand-object interactions, and long-tailed class distributions. Clip length in EK100 varies widely. To maximize temporal coverage without using test-time augmentations, we sample frames video-length-dependent. We choose smallest stride such that we are not missing more than $2\times\text{stride}$ each left and right of the temporally-centered crop. We use a maximal stride of 20.

\paragraph{Something-Something V2 (SSv2) \cite{goyal_something_2017}.}
Something-Something V2 contains short clips of people performing object manipulation actions in front of a static camera, annotated with 174 templated action classes such as ``moving something from left to right'' or ``pretending to put something into something''. Because the templates abstract away from specific objects and scenes, recognition on SSv2 strongly depends on the temporal dynamics of the interaction rather than on appearance.

\paragraph{Invariant Action Recognition Dataset (IARD) \cite{tacchetti_invariant_2017}.}
The Invariant Action Recognition Dataset contains short clips of actors performing a controlled set of actions under systematic variations in viewpoint and actor identity. The dataset was originally introduced to study invariant action recognition in human visual cortex and supports controlled evaluation of how robust learned representations are to such nuisance transformations. We use it to assess the viewpoint and actor invariance.

\paragraph{Chirality in Action (CiA) \cite{bagad_chirality_2026}.}
The Chirality in Action benchmark is a curated subset of SSv2, EK100 and Charades that focuses on chiral action pairs, i.e., actions whose meaning changes when the video is played in reverse or mirrored (e.g., ``moving something from left to right'' vs. ``moving something from right to left''). The benchmark is specifically designed to probe whether models capture the temporal direction and asymmetry of motion, rather than relying on appearance shortcuts, and is therefore well-suited to evaluating trajectory-based representations. In this work, we evaluate on the SSv2 split from Chirality in Action.

\begin{table}[htb]
    \centering
    \footnotesize
    \caption{\textbf{Overview of the datasets used in our experiments.}
             Numbers correspond to the splits and label spaces used in this
             paper. For EK100 we evaluate on the verb label space, and since
             test labels are not publicly available, evaluation is performed
             on the validation set. For IARD, we report statistics for both
             the Action and ID classification splits.}
    \label{tab:datasets}
    \renewcommand{\arraystretch}{1.15}
    \begin{tabularx}{\textwidth}{l C C C C C C}
        \toprule
        & & & & \multicolumn{2}{c}{\textbf{IARD}} & \\
        \cmidrule(lr){5-6}
        & \textbf{K400} & \textbf{EK100} & \textbf{SSv2} & \textbf{Action} & \textbf{Identity} & \textbf{CiA-SSv2} \\
        \midrule
        \textbf{Domain}         & Web videos       & Egocentric cooking & Object interactions & \multicolumn{2}{c}{Controlled lab}                & Object interactions \\
        \textbf{Viewpoint}      & exo-centric      & ego-centric        & exo-centric        & \multicolumn{2}{c}{exo-centric}        & exo-centric \\
        \textbf{\#Classes}      & 400              & 97 (verbs)         & 174                 & \multicolumn{2}{c}{5}                 & 2 (per group) \\
        \textbf{\#Train clips}  & 241{,}181  & 67{,}217           & 168{,}913           & 3{,}720             & 3{,}712         & 12{,}216 \\
        \textbf{\#Val clips}    & 19{,}877   & 9{,}668            & 24{,}777            & --                  & --              & 1{,}430 \\
        \textbf{\#Test clips}   & 38{,}671   & 13{,}092           & 27{,}157            & 920                 & 928             & -- \\
        \textbf{Avg.\ frames}   & 262              & 173                & 45.8                & \multicolumn{2}{c}{60.0}              & 43.2 \\
        \textbf{Avg.\ length (s)} & 9.6            & 3.1                & 3.8                 & \multicolumn{2}{c}{2.0}               & 3.6 \\
        \textbf{Resolution}     & Variable         & 1080p              & 240p                & \multicolumn{2}{c}{600$\times$1080}                & 240p \\
        \textbf{Source}         & YouTube          & Head-mounted cam   & Crowd-sourced       & \multicolumn{2}{c}{Lab recordings}    & Subset of SSv2 \\
        \bottomrule
    \end{tabularx}
\end{table}

\section{Augmentations}
\label{app:augments}
To improve the generalization of our trajectory model, we apply a sequence of stochastic augmentations during training. All augmentations operate directly on the extracted point trajectories and are designed to be consistent with the geometric structure of trajectory data, i.e., they preserve the relative motion patterns that our model is intended to recognize. We apply the following augmentations:

\paragraph{Global translation.}
We translate all trajectories in a clip jointly by a random offset $(\Delta x, \Delta y)$ in the image plane, with $\Delta x$ and $\Delta y$ sampled uniformly from a fixed range. This encourages invariance to the absolute position of the action in the frame.

\paragraph{Global rotation.}
We apply a random rotation around the $z$-axis (i.e., in the image plane) to all trajectories of a clip jointly, with the rotation angle sampled uniformly from $[-\theta_{\max}, \theta_{\max}]$. This makes the model more robust to small variations in camera roll and viewpoint orientation while preserving the relative motion between trajectories.

\paragraph{Temporal shifting with padding.}
We randomly shift the trajectories along the temporal axis by an offset sampled uniformly from a fixed range, padding the resulting empty frames at the beginning or end of the clip. This augmentation discourages the model from relying on the absolute timing of an action within the input window.

\paragraph{Trajectory jitter.}
Independently for each trajectory, we add small Gaussian noise to the per-frame $(x, y)$ coordinates. Unlike the global augmentations above, this noise is applied per-trajectory rather than jointly, which simulates imperfect trajectory extraction and prevents the model from overfitting to exact trajectory geometry.

\paragraph{Spatial cropping.}
We randomly crop a region in the $(x, y)$ plane and discard trajectories that fall outside it; the remaining trajectories are recentered into the crop. This acts analogously to random spatial cropping in image-based training and exposes the model to different spatial extents of the underlying scene.

\section{Detailed Experimental Setup}
\label{app:detail_exp}
This section provides the detailed experimental setup for each of the experiments reported in the main paper. Hyperparameter values are listed in Appendix~\ref{app:hyperparams}.

\paragraph{Pretraining.}
We pretrain our trajectory model on the training split of the respective target dataset only. Pretraining operates on trajectory sequences of $32$ timesteps, with coordinates expressed in normalized image coordinates. On top of the trajectory encoder, we attach a lightweight bottleneck followed by a decoder consisting of two self-attention layers and a final prediction MLP. The decoder is trained to reconstruct the input trajectory coordinates, and we optimize the mean absolute error (L1) between the reconstructed and ground-truth normalized coordinates. After pretraining, the decoder is discarded and only the encoder weights are retained for downstream training.

\paragraph{Single model evaluation.}
For each main dataset, we fine-tune our trajectory model on the corresponding training split only. On top of the pretrained encoder, we attach a classification head consisting of a linear layer that projects the \texttt{[CLS]} token to the number of action classes of the dataset. The model is trained end-to-end using the standard cross-entropy loss.

\paragraph{Vision baseline evaluation.}
To compare against image and video foundation models on equal footing, we re-evaluate all vision baselines under a unified protocol rather than relying on the heterogeneous evaluation setups used in the respective original papers. Specifically, for every vision backbone we (i) extract and cache features once, using a center crop at the model's native input resolution and a single clip with a single crop (i.e., the standard $16{\times}1{\times}1$ protocol), and (ii) train an attentive probe on top of the cached features following the single cross-attention probe design from V-JEPA~\cite{bardes_revisiting_2024}. For image models, we add a $3$D positional encoding on top of the cached features to provide explicit temporal information. We cache features to save compute and under the assumption that well-trained vision foundation encoders are largely robust to simple augmentations. We use a test-time augmentation free setting for all evaluated models including our own, to ensure a fair comparison between methods. This protocol ensures that differences between methods reflect the underlying representations rather than differences in test-time augmentation or probe design.

\paragraph{I3D baseline}
As optical flow baseline, we take only the flow-branch of I3D~\cite{carreira_quo_2017}. I3D processed RGB and flow features separately and only adds logits, which makes it possible to use the flow-branch standalone. We use a checkpoint that is pretrained on both ImageNet and Kinetics-400 (supervised pretraining). We extract optical flow using RAFT-Small. We extract flow once at fixed temporal locations with frames resized to 256 short side, clipping flow values larger than 20 pixels. Following previous work, we extract 16 frames via linspace sampling on SSv2 and 16 frames with stride 4 on K400. For EK100, we use the same adaptive sampling strategy employed for our model. This increases temporal coverage compared to the standard stride 4 sampling, which is important as we do not use test-time augmentations. During training, we use random cropping and random horizontal flip as augmentations and train with $p=0.5$ dropout. We use SGD with 0.9 momentum and an initial learning rate of 0.02. Following~\cite{carreira_quo_2017}, we use a step learning rate scheduler, reducing to 0.002 after 8 epochs for K400 and SSv2 and 15 epochs for the smaller EK100. 

\paragraph{Complementary fusion experiments.}
For the complementary experiments combining our trajectory model with vision backbones, we follow the protocol introduced in the Chirality in Action paper~\cite{bagad_chirality_2026}. Both the trajectory model and the attentive probe on top of the vision backbone are first trained on the full training set of the respective dataset. We then select the best model checkpoint based on validation accuracy (or the last checkpoint on EK100), discard the final linear classification layer of each model, and freeze both backbones. A new linear classifier is then trained on top of the concatenated features. This protocol isolates the complementarity of the two representations.

\paragraph{IARD.}
For the Invariant Action Recognition Dataset (IARD)~\cite{tacchetti_invariant_2017}, we follow the evaluation protocol of DisMo~\cite{ressler-antal_dismo_2026}. The dataset is evaluated under two splits: an \emph{action} split, in which trajectories of four participants are used for training and a held-out fifth participant is used for testing; and an \emph{identity} split, in which four actions are used for training and a held-out fifth action is used for testing. Evaluation is performed by $k$-NN classification on mean-pooled backbone features with $k = 20$: we cache the pooled embeddings for all test clips, and predictions on the validation set are obtained via weighted majority voting over the $k$ nearest training embeddings. Baseline results for V-JEPA, VideoMAE, and DisMo are taken directly from the DisMo paper. DINOv2 and our trajectory model are evaluated by us under the same protocol.

\paragraph{Chirality in Action (CiA-SSv2).}
For Chirality in Action, we follow the standard protocol of \cite{bagad_chirality_2026}. For each chiral group, we train a linear classifier on top of the backbone features using the training set of that group. We report the average accuracy across all chiral groups on the test set.

\section{Hyperparameters}
\label{app:hyperparams}
The hyperparameters used during pretraining and finetuning are listed in \autoref{tab:hyperparams}. We used the same parameters for all datasets. In the complementary experiments, we concatenate the trajectory model's cls token with the vision embedding after the attentive pooling and train a single linear layer on top. Therefore, the model dimensions are fixed. We train the linear layer using the same setup as for finetuning, except that we reduce the epochs to 5, increase the weight decay to $5e-2$ and increase the warmup ratio to 0.4. For the CiA-SSv2 experiments, we use the same hyperparameter as for finetuning, except that we decrease the learning rate to $1e-4$ and reduce the maximum epochs to 10, while performing at least 100 training iterations. This is done to guarantee that the linear readout is trained long enough on small groups. 

\begin{table}[htb]
    \centering
    \caption{\textbf{Hyperparameters used for pretraining and fine-tuning of our trajectory model.} \emph{Pretraining} refers to the self-supervised reconstruction stage on the respective training split. \emph{Fine-tuning} refers to the supervised classification stage on the same dataset. Entries marked -- are not applicable to the corresponding stage.}
    \label{tab:hyperparams}
    \renewcommand{\arraystretch}{1.15}
    \begin{tabularx}{\textwidth}{l C C}
        \toprule
        & \textbf{Pretraining} & \textbf{Fine-tuning} \\
        \midrule
        \multicolumn{3}{@{}l}{\textit{Optimization}} \\
        Epochs              & 300   & 15 \\
        Batch size          & 512   & 64 \\
        Optimizer           & adamW~\cite{loshchilov_decoupled_2019} & adamW \\
        Learning rate       & 2e-4 & 2e-4 \\
        $\beta_1$           & 0.9  & 0.9 \\
        $\beta_2$           & 0.999  & 0.999 \\
        Weight decay        & 5e-3  & 5e-3 \\
        LR scheduler        & cosine  & cosine \\
        Warmup ratio        & 0.1  & 0.1 \\
        \midrule
        \multicolumn{3}{@{}l}{\textit{Trajectory input}} \\
        \#Trajectories per clip & 512 & 512 \\
        Trajectory length (frames) & 32 & 32 \\
        \midrule
        \multicolumn{3}{@{}l}{\textit{Encoder architecture}} \\
        Embedding dim       & 288 & 288 \\
        \#Layers            & 6 & 6 \\
        \#Heads             & 4 & 4 \\
        Sinusoidal embedding dim & 32 & 32 \\
        \midrule
        \multicolumn{3}{@{}l}{\textit{Decoder architecture (pretraining only)}} \\
        Mask ratio          & 0.9 & -- \\
        Decoder embedding dim & 144 & -- \\
        Decoder \#layers    & 2 & -- \\
        Decoder \#heads     & 4 & -- \\
        \bottomrule
    \end{tabularx}
\end{table}

\section{Trajectory extraction}
\label{app:extraction}
All trajectories are extracted in float16 using \texttt{torch.amp}. The temporal sampling differs slightly per dataset to match the effective time horizon used by the vision-model baselines and to accommodate variable-length videos. On Something-Something v2 and IARD we extract all available frames and interpolate the resulting trajectories to a fixed length of 32 frames. On Kinetics-400 we extract 32 frames at a stride of 2, matching the temporal horizon of the vision baselines, which use 16 frames at stride 4. On EPIC-Kitchens-100, where action segments vary substantially in length, we extract 32 trajectories with a stride that depends on the segment duration.

\section{Visualizations}
\label{app:vis}
\begin{figure*}[htb]
    \centering
    \begin{subfigure}{\textwidth}
        \centering
        \includegraphics[width=\textwidth]{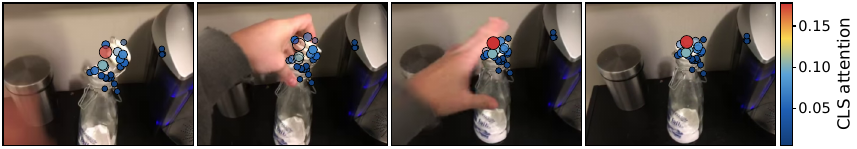}
        \label{fig:overview_a}
    \end{subfigure}

    \begin{subfigure}{\textwidth}
        \centering
        \includegraphics[width=\textwidth]{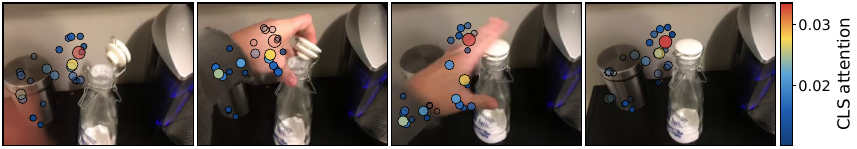}
        \label{fig:overview_b}
    \end{subfigure}

    \begin{subfigure}{\textwidth}
        \centering
        \includegraphics[width=\textwidth]{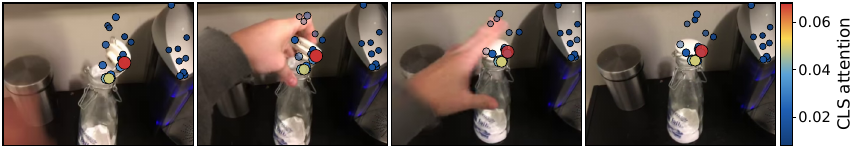}
        \label{fig:overview_c}
    \end{subfigure}

    \begin{subfigure}{\textwidth}
        \centering
        \includegraphics[width=\textwidth]{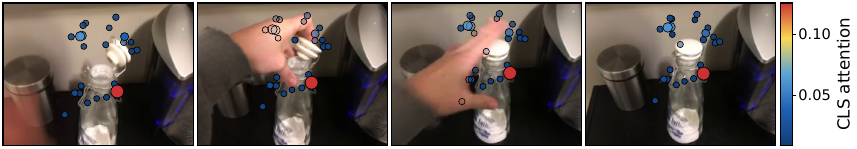}
        \label{fig:overview_d}
    \end{subfigure}

    \caption{\textbf{Last-layer CLS attention overlay over all heads.} Different heads focus on different trajectories. Head 1 and 3 focus on the trajectories covering the bottle cap, whereas head 2 and 4 focus on the head movement either directly through trajectories on the hand or through background trajectories which get occluded.}
    \label{fig:attention_vis_ext}
\end{figure*}

We visualize the last-layer CLS attention of all four attention heads in \autoref{fig:attention_vis_ext}. It can be seen that different heads attend to different sets of trajectories.

\section{Computational resources}
\label{app:resources}
The full pipeline involves three stages with distinct compute requirements. Trajectory extraction (including depth estimation) across all datasets totals approximately 360 A100 GPU-hours and is performed once. Pretraining the trajectory model on Kinetics-400 takes around 12 A100 GPU-hours per run, and fine-tuning on Kinetics-400 takes approximately 0.75 A100 GPU-hours per run. Trajectory and depth features are precomputed and cached, so neither contributes to the per-run cost of training or fine-tuning. We estimate the total compute for this project, including ablations, hyperparameter exploration, and unreported experiments, at approximately 900 A100 GPU-hours.

\section{Broader Impacts}
\label{app:broader_impacts}
Our work advances action recognition from point trajectories, with potential positive impact in assistive technology, human-robot interaction, and video accessibility. Operating on sparse trajectories rather than raw pixels also reduces the amount of identifying visual information processed by the model.

As with most human-centric vision research, action recognition has an indirect connection to surveillance and could be misused in ways that raise privacy or fairness concerns. Our contribution is methodological: we train closed-set classifiers on established public benchmarks (SSv2, EK100, K400), and the resulting model does not identify individuals or generate content. Its capabilities are not qualitatively different from those of existing action recognition systems. As with any such model, predictions can be incorrect and performance may vary across demographic groups, so we caution against deployment in high-stakes settings without application-specific evaluation.

\section{Licenses}
\label{app:licenses}

\subsection{Licenses of datasets used}

We list below the licenses, terms of use, and sources of all datasets used in
this paper. We respect the terms of use of each dataset and use them strictly
for non-commercial academic research.

\begin{itemize}
    \item \textbf{Kinetics-400 (K400)}~\cite{carreira_quo_2017}\newline
          License: not specified by the original release. Videos are sourced from YouTube and remain subject to the terms of their original uploaders. We use the dataset for non-commercial academic research.\\
          Source: \url{https://github.com/cvdfoundation/kinetics-dataset}.
    \item \textbf{EPIC-Kitchens-100 (EK100)}~\cite{damen_rescaling_2022}\newline
          License: Creative Commons Attribution-NonCommercial 4.0 International (CC BY-NC 4.0).\\
          Source: \url{https://epic-kitchens.github.io/2026}.
    \item \textbf{Something-Something V2 (SSv2)}~\cite{goyal_something_2017}\newline
          License: Qualcomm Research Use License.\\
          Source: \url{https://www.qualcomm.com/developer/software/something-something-v-2-dataset}.
    \item \textbf{Invariant Action Recognition Dataset (IARD)}~\cite{tacchetti_invariant_2017}\newline
          License: Creative Commons Zero v1.0 Universal (CC0 1.0).\\
          Source: \url{https://dataverse.harvard.edu/dataset.xhtml?persistentId=doi:10.7910/DVN/DMT0PG}.
    \item \textbf{Chirality in Action (CiA-SSv2)}~\cite{bagad_chirality_2026}\newline
          A curated subset of SSv2. The underlying videos are governed by the SSv2 license above.\\
          Source: \url{https://github.com/bpiyush/LiFT}.
\end{itemize}

\subsection{Licenses of pretrained models used}

We additionally use the following pretrained vision foundation models as baselines and/or feature extractors. We respect the terms of use of each model and use them strictly for non-commercial academic research.

\begin{itemize}
    \item \textbf{CoTracker3}~\cite{karaev_cotracker3_2025}\newline
          License: Creative Commons Attribution-NonCommercial 4.0 International (CC-BY-NC 4.0).\\
          Source: \url{https://github.com/facebookresearch/co-tracker}.
    \item \textbf{Video Depth Anything}~\cite{chen_video_2025}\newline
          License: Creative Commons Attribution-NonCommercial 4.0 International (CC-BY-NC 4.0).\\
          Source: \url{https://github.com/DepthAnything/Video-Depth-Anything}.
    \item \textbf{DINOv2}~\cite{oquab_dinov2_2024}\newline
          License: FAIR Noncommercial Research License.\\
          Source: \url{https://github.com/facebookresearch/dinov2}.
    \item \textbf{SigLIP 2}~\cite{tschannen_siglip_2025}\newline
          License: Apache License 2.0 (code) and Creative Commons Attribution 4.0 International (CC-BY 4.0, model weights).\\
          Source: \url{https://github.com/google-research/big_vision/blob/main/big_vision/configs/proj/image_text/README_siglip2.md}.
    \item \textbf{VideoMAE}~\cite{tong_videomae_2022}\newline
          License: Creative Commons Attribution-NonCommercial 4.0 International (CC-BY-NC 4.0).\\
          Source: \url{https://github.com/mcg-nju/videomae}.
    \item \textbf{VideoPrism}~\cite{zhao_videoprism_2024}\newline
          License: Apache License 2.0 (code) and Creative Commons Attribution 4.0 International (CC-BY 4.0, model weights).\\
          Source: \url{https://github.com/google-deepmind/videoprism}.
    \item \textbf{V-JEPA 2}~\cite{assran_vjepa_2025}\newline
          License: MIT License (with portions of the training code under Apache License 2.0).\\
          Source: \url{https://github.com/facebookresearch/vjepa2}.
    
\end{itemize}

\newpage

\end{document}